\documentclass{article}

\usepackage{bm}
\usepackage{amsfonts}
\usepackage{dsfont}
\usepackage{amsmath}
\usepackage{subfig}
\usepackage{graphicx}
\usepackage{floatrow}
\usepackage[ruled,linesnumbered]{algorithm2e}
\usepackage[square,numbers]{natbib}
\newfloatcommand{capbtabbox}{table}[][\FBwidth]
\newsavebox{\measurebox}

     \PassOptionsToPackage{numbers, compress, square}{natbib}

\usepackage[preprint]{neurips_2020}

\usepackage[utf8]{inputenc} 
\usepackage[T1]{fontenc}    
\usepackage{hyperref}       
\usepackage{url}            
\usepackage{booktabs}       
\usepackage{amsfonts}       
\usepackage{nicefrac}       
\usepackage{microtype}      

\newcommand{\hsp}{\hspace{4mm}}

\newcommand{\hsppp}{\hspace{12mm}}

\newcounter{commentcounter}
\long\def\symbolfootnote[#1]#2{\begingroup%
\def\thefootnote{\fnsymbol{footnote}}\footnote[#1]{#2}\endgroup}

\newcommand{\DL}{\mathds{L}}

\title{Rule Covering for Interpretation and Boosting}

\author{
    Ş. İlker Birbil\thanks{Corresponding author.} \\
    Econometric Institute \\
    Erasmus University Rotterdam \\
    3000 DR Rotterdam, The Netherlands\\
    \texttt{birbil@ese.eur.nl} \\
\And
    Mert Edali \\
    Department of Medicine \\ 
    University of Chicago \\
    Chicago, IL 60637, USA \\
    \texttt{meadli@medicine.bsd.uchicago.edu} \\
\And
    Birol Yüceoğlu \\
    Migros T.A.Ş \\ Ataşehir, 34758, İstanbul, Turkey \\
    \texttt{biroly@migros.com.tr} \\
}

\begin{document}

\maketitle

\begin{abstract}
  We propose two algorithms for interpretation and boosting of
  tree-based ensemble methods. Both algorithms make use of mathematical
  programming models that are constructed with a set of rules
  extracted from an ensemble of decision trees. The objective is to
  obtain the minimum total impurity with the least number of rules
  that cover all the samples. The first algorithm uses the collection
  of decision trees obtained from a trained random forest model. Our
  numerical results show that the proposed rule covering approach
  selects only a few rules that could be used for interpreting the
  random forest model. Moreover, the resulting set of rules closely
  matches the accuracy level of the random forest model. Inspired by
  the column generation algorithm in linear programming, our second
  algorithm uses a rule generation scheme for boosting decision
  trees. We use the dual optimal solutions of the linear programming
  models as sample weights to obtain only those rules that would
  improve the accuracy. With a computational study, we observe that
  our second algorithm performs competitively with the other
  well-known boosting methods. Our implementations also demonstrate
  that both algorithms can be trivially coupled with the existing
  random forest and decision tree packages.
\end{abstract}

\section{Introduction}
\label{sec:intro}

Although their prediction performances are remarkable, tree-based
ensemble methods are difficult to interpret due to large number of
trees in the trained model. Among these methods, Random Forest
algorithm is probably the most frequently used alternative for solving
classification and regression problems arising in different domains
\cite{breiman2001random}. This performance versus interpretability
trade-off leads to the following two main questions in our work: Given
a trained random forest, can we obtain an interpretable representation
with mathematical programming that shows a performance close-enough to
the overall forest? Using a similar mathematical programming model,
can we come up with a learning algorithm, which generates only those
rules that improve the performance of a base tree?

To answer these questions, we focus on the rules constituting the
trees in the forest and propose two algorithms based on mathematical
programming models. The model in the first algorithm extracts the
rules corresponding to the leaves of the forest that result in the
minimum total impurity while covering all the samples. With this
algorithm, we are able to mimick the performance of the original
forest with a significantly few number of rules (most of the time even
less than the number of rules in a regular decision tree). Being
encouraged with these results, we then propose our second algorithm
based on selective generation of rules. This algorithm improves the
total impurity gradually by solving a series of linear programming
problems. This approach is a variant of the column generation
algorithm in mathematical optimization. In our scheme, we use the dual
optimal solutions of the linear programs as the sample weights. We
also observe that this scheme has close ties to well-known ideas in
boosting methods \citep{mason99, mason00, schapire98}. Thus, rule
extraction and boosting are the two keywords that help us to review
the related literature.

\citet{friedman2008predictive} propose RuleFit method to extract rules
from ensembles. The authors use decision trees as base learners, where
\textit{each node} of the decision tree corresponds to a rule. They
present an ensemble generation algorithm to obtain a rule set. Then,
they solve a linear regression problem that minimizes the loss with a
lasso penalty for selecting rules. \citet{meinshausen2010node}
presents a similar approach, called Node Harvest, which relies on
rules generated by a tree ensemble. Node Harvest minimizes the
quadratic loss function by partitioning the samples and relaxing the
integrality constraints on the weights. Even though the algorithm does
not constrain the number of rules, the author observes that only a
small number of rules are used. \citet{mashayekhi15} propose coverage
of the samples in the training set by the rules generated by Random
Forest algorithm. They create a score for each rule by considering the
percentage of correctly and incorrectly classified samples as well as
the rule length. By using a hill climbing algorithm, they are able to
obtain slightly lower accuracy with much fewer rules than random
forest. Extending this work, \citet{mashayekhi2017rule} propose three
different algorithms to extract \texttt{if-then} rules from decision
tree ensembles. The authors then compare their algorithms with both
RuleFit and Node Harvest. All these approaches use accuracy values to
construct the objective functions. Interpretability in terms of
reducing the number of rules is achieved by using lasso penalties. Our
first algorithm, on the other hand, reduces the number of rules
directly through the objective function.

Gradient boosting algorithms use a sequence of weak learners and build
a model iteratively \cite{freund97, mason99, friedman2001greedy}. At
each iteration the boosting algorithm focuses on the misclassified
samples by adjusting their weights with respect to the errors of the
previous iterations. A boosting method that uses linear programming is
called LPBoost \cite{demiriz2002linear}. This algorithm uses weak
learners to maximize the margins separating the different
classes. Like us, the authors also use column generation to create the
weak learners iteratively. The final ensemble then becomes a weighted
combination of the weak learners. The approach can accept decision
stumps or decision trees as weak learners. However, decision trees can
perform badly in terms of convergence and solution accuracy, if the
dual variables associated with margin classification are not
constrained. Unlike LPBoost, we use the duals as sample weights in our
column generation routine. Another boosting method based on
mathematical programming is called IPBoost
\citep{pfetsch2020ipboost}. This approach uses binary decision
variables corresponding to misclassification by a predetermined
margin. Due to binary decision variables, the authors resort to an
integer programming solution technique called
branch-and-bound-and-price. \citet{dash18} also use column generation
to learn Boolean decision rules in either disjunctive normal form or
conjunctive normal form. They solve a sequence of integer programming
problems for pricing the columns. They choose to restrict the number
of selected clauses as a user-defined hyperparameter for improving
interpretability. \citet{bertsimas2017optimal} propose an integer
programming formulation to create optimal decision trees for
multi-class classification. They present two approaches with respect
to node splitting. The first approach is more general but less
interpretable. The second more interpretable model considers the
values of one feature for node splitting. One advantage of their model
is that it does not rely on an assumption for the nature of the values
of the features. \citet{gunluk2018optimal} propose a binary
classification tree formulation for only categorical features. The
idea can be extended to numerical variables by using thresholding. The
approach is less general than the approach presented by
\citet{bertsimas2017optimal}, but has some advantages like the number
of integer variables being independent of the training dataset
size. \citet{firat20} propose a column generation based heuristic for
learning decision tree. The approach is based on generating decision
paths and can be used for solving instances with tens of thousands of
observations.

In the light of this review, we make the following contributions to
the literature: We propose a new mathematical programming approach for
interpretation and boosting of random forests and decision trees,
respectively. Unlike other work in the literature, the objective
function in our models aims at minimizing both the total impurity and
the number of selected rules. When applied to a trained random forest
for interpretation, our first algorithm obtains significantly few
number of rules with an accuracy level close to the random forest
model. Our second algorithm is a new boosting method based on rule
generation and linear programming. As a novel approach, the algorithm
uses dual information to weigh samples in order to increase the
coverage with less impure rules. In our computational study, we
demonstrate that both algorithms are remarkably easy to implement
within a widely-used machine learning package.\footnote{(GitHub page)
  -- \url{https://github.com/sibirbil/RuleCovering}} Without any fine
tuning, we obtain quite promising numerical results with both
algorithms showing the potential of the rule covering for
interpretation and boosting.

\section{Minimum rule cover: interpretation of forests}
\label{sec:mirco}

Let $(\bm{x}_i, y_i)$, $i \in \mathcal{I}$ be the set of samples,
where the vector $\bm{x}_i \in \mathcal{X}$ and the scalar
$y_i \in \mathcal{K}$ denote the input sample and the output class,
respectively. We shall assume in the subsequent part that the learning
problem at hand is a classification problem. However, our discussion
here can be extended to regression problems as well (we elaborate on
this point in Section \ref{sec:extensions}.\footnote{All cross
  references starting with letter ``S'' refer to the supplementary
  document.}).

Suppose that a Random Forest algorithm is trained on this dataset and
a collection of trees is grown. Given one of these trees, we can
easily generate the path $j$ from the \textit{root node} to a
\textit{leaf node} that results in a subset of samples $\bm{x}_i$,
$i \in \mathcal{I}(j) \subseteq \mathcal{I}$. Actually, these paths
constitute the rules that are later used for classification with
majority voting. Each rule corresponds to a sequence of
\texttt{if-then} clauses and it is likely that some of these sequences
may appear multiple times. We denote all such rules in all trees by
set $\mathcal{J}$. Clearly, the size of $\mathcal{J}$ is quite
large. Therefore, we next construct a mathematical programming model
that aims at selecting the minimum number of rules from the trained
forest while preserving the performance. We also make sure that all
samples are covered with the selected rules.

As rule $j \in \mathcal{J}$ corresponds to a leaf node, we can also
evaluate the node impurity, $w_j$. Two most common measures for
evaluating the node impurity are known as \textit{Gini} and
\textit{Entropy} criteria. We further introduce the binary decision
variables $z_j$ that mark whether the corresponding rules
$j \in \mathcal{J}$ are selected or not. The cost of $w_j$ is
incurred, when $z_j=1$. Note that if we use only these impurity
evaluations as the objective function coefficients (costs), then the
model would tend to select many rules that cover only a few samples,
which are likely to be \emph{pure nodes}. However, we want to advocate
the use of fewer rules in the model so that the resulting set of rules
is easy to interpret. In other words, we also aim at minimizing the
number of rules along with the impurity cost. Therefore, we replace
the cost coefficient so that selecting excessive number of rules is
penalized. Our mathematical programming model then becomes:
\begin{equation}
  \label{eq:scmodel}
  \begin{array}{lll}
    \mbox{minimize\hspace{4mm}} & \sum_{j \in \mathcal{J}} (1+w_j) z_j & \\[2mm]
    \mbox{subject to\hspace{4mm}} & \sum_{j \in \mathcal{J}(i)} z_j \geq 1, & i \in \mathcal{I}, \\[2mm]
              & z_j \in \{0,1\}, & j \in \mathcal{J},
  \end{array}
\end{equation}
where $\mathcal{J}(i)$, $i \in \mathcal{I}$ is the set of rules
prescribing all the leaves involving sample $\bm{x}_i$. In other
words, sample $i$ is covered with the rules in subset
$\mathcal{J}(i) \subseteq \mathcal{J}$.

The mathematical programming model \eqref{eq:scmodel} is a standard
weighted set covering formulation, where sets and items correspond to
rules and samples, respectively. Although set covering problem is
NP-hard, current off-the-shelf solvers are capable of providing
optimal solutions to moderately large instances. In case the integer
programming formulation becomes difficult to solve, there are also
powerful heuristics that return approximate solutions very
quickly. One such approach is the well-known greedy algorithm proposed
by Chvatál \cite{chvatal1979greedy}. Algorithm \ref{alg:greedy} shows
the steps of this heuristic using our notation. Recently, Chvatál's
greedy algorithm, as we implement here, has been shown to perform
quite well on a set of test problems when compared against several
more recent heuristics \cite{vasko16}. Thus, we have reported our
results also with this algorithm in Section
\ref{sec:compstudy}. Clearly, this particular heuristic can be
replaced with any other approach solving the set covering
problem. Since these heuristics provide an approximate solution, we
denote the resulting set of rules by $\hat{\mathcal{J}}$, and by
definition, the cardinality of set $\hat{\mathcal{J}}$ is larger than
the cardinality of the optimal set of rules obtained by solving
\eqref{eq:scmodel}.

Algorithm \ref{alg:mirco} gives the steps of our minimum rule cover
algorithm (MIRCO). The algorithm takes as input the trained random
forest model and the training dataset. Note that after running
Algorithm \ref{alg:mirco}, we obtain a subset of rules
$\hat{\mathcal{J}}$ that can also be used for classifying
out-of-sample points (test set) with majority voting.  If we denote
the predicted class of test sample $\bm{x}_0$ with
$C(\bm{x}_0, \hat{\mathcal{J}})$, then
\begin{equation}
  \label{eq:predict}
  C(\bm{x}_0, \hat{\mathcal{J}}) = \arg\max_{k \in \mathcal{K}}\big\{\sum_{j \in \hat{\mathcal{J}}}
    n_{jk}\DL_j(\bm{x}_0)\big\}, 
\end{equation}
where $n_{jk}$ stands for the number of samples from class $k$ in leaf
$j$, and $\DL_j(\bm{x}_0)$ is an operator showing whether $\bm{x}_0$
satisfies the sequence of clauses corresponding to leaf $j$. In fact,
testing in this manner would be a natural approach for evaluating the
performance of MIRCO. Since our main objective is to interpret the
underlying model obtained with Random Forest algorithm, we hope that
this subset of rules would obtain a classification performance closer
to the performance obtained with all the rules in set
$\mathcal{J}$. We show with our computational study in Section
\ref{sec:compstudy} that this is indeed the case.

\begin{algorithm}
  \SetAlgoLined
  \SetKwInOut{Input}{Input}
  \Input{Random forest model \texttt{RF}; training dataset $(\bm{x}_i, y_i)$, $i \in \mathcal{I}$}
  $\mathcal{J} = \emptyset$\;
  \For{\texttt{\upshape Tree} \KwSty{in} \texttt{\upshape RF}}{
    $\mathcal{J} \leftarrow \mathcal{J} \cup \text{(rules in \texttt{Tree})}$\;
  }

  Evaluate impurities $w_j$, $j \in \mathcal{J}$ using $(\bm{x}_i, y_i)$,
  $i \in \mathcal{I}$\; \label{ln:ei}

  Construct sets $\mathcal{J}(i)$, $i \in \mathcal{I}$ \;

  Solve model \eqref{eq:scmodel} with Algorithm
  \ref{alg:greedy}\;

  Return $\hat{\mathcal{J}}$\;
\caption{MIRCO}
\label{alg:mirco}
\end{algorithm}

\begin{figure}
    \centering
    \includegraphics[scale = 0.6, trim={0cm, 0.5cm, 0cm, 0cm}]{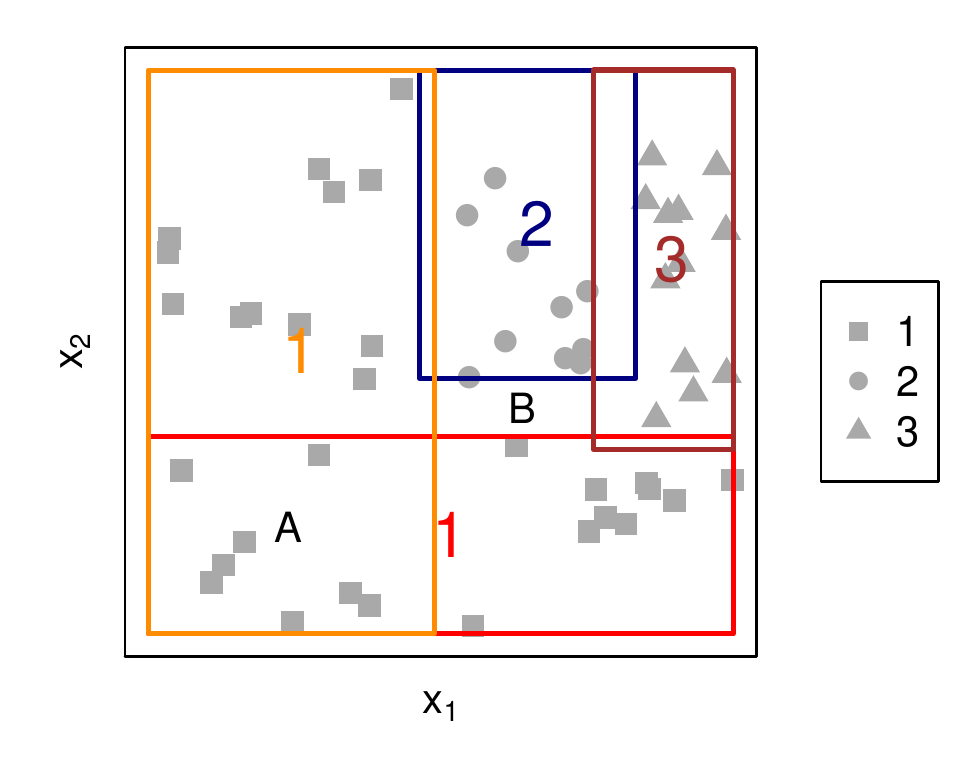}
    \caption{A toy example illustrating that MIRCO does not
      necessarily correspond to a tree. Rectangles denote the regions
      obtained with the rules in $\hat{\mathcal{J}}$. The numbers in
      the rectangles depict the classification results corresponding to
      three classes shown with three different markers.}
    \label{fig:misses}
\end{figure}

Figure \ref{fig:misses} shows an example set of of rules obtained with
MIRCO on a small data set. Clearly, the resulting set of rules does
not correspond to a decision tree. On one hand, this implies that we
cannot simply select one of the trees grown by the Random Forest
algorithm and replace its leaves with the set of rules obtained with
MIRCO. Consequently, multiple rules may cover the same test sample, in
which case we can again use majority voting to determine its
class. Region A in Figure \ref{fig:misses} shows that any test sample
in that region will be classified with two overlapping rules. On the
other hand, there is also the risk that the entire feature space is
not covered by the subset of rules. In Figure \ref{fig:misses} such a
region is marked as B, where none of the rules cover this
region. Therefore, if one decides to use only the rules in
$\hat{\mathcal{J}}$ as a \textit{classifier}, then some of the samples
in the \textit{test set} may not be classified at all. This is an
anticipated behaviour, since MIRCO guarantees to cover only those
samples that are used to train the Random Forest algorithm. Our
numerical results show that the percentages of test samples that MIRCO
fails to classify are quite low.

Notice that the entire set of rules used by MIRCO does not have to be
constructed after training a Random Forest algorithm. If there is an
oracle that provides many rules to construct the set $\mathcal{J}$,
then we can still (approximately) solve the mathematical programming
model \eqref{eq:scmodel} and obtain the resulting set of rules. In
fact, the development of this oracle decouples our discussion from
training a forest. In the next section, we propose such an oracle that
generates the necessary rules on the fly.

\section{RCBoost: rule cover boosting}
\label{sec:secoma}

The main difficulty with the set covering problem is the integrality
condition on the decision variables. When this condition is ignored,
the resulting problem is a simple linear programming problem that can
be solved quite efficiently. For the same reason, many approximation
methods devised for the set covering problem are based on considering
the linear programming relaxation of \eqref{eq:scmodel}. Formally, the
relaxed problem simply becomes:
\begin{equation}
  \label{eq:scappmodel}
  \begin{array}{lllc}
    \mbox{minimize\hspace{4mm}} & \sum_{j \in \mathcal{J}} (1+w_j) z_j & &\\[2mm]
    \mbox{subject to\hspace{4mm}} & \sum_{j \in \mathcal{J}(i)} z_j \geq 1, &
                                         i \in \mathcal{I}, \hsppp (\lambda_i) \\[2mm]
                                & z_j \geq 0, & j \in \mathcal{J}, &
  \end{array}
\end{equation}
where $\lambda_i$, $i \in \mathcal{I}$ denote the dual variables
corresponding to the constraints. After solving this problem, we
obtain both the primal and the dual optimal solutions. As we will
discuss shortly, the optimal dual variables bear important information
about the sample points. Since we relax the integrality restrictions,
one suggestion could be rewriting the objective as solely the
minimization of total impurity. We refrain from such an approach
because when there are many pure leaves, then the corresponding rule
does not contribute to the objective function value ($w_j = 0$) and
the corresponding $z_j$ variable becomes free to attain any
value. This implies that the primal problem has multiple optimal
solutions, and hence causing degeneracy in the dual.

We have seen in the previous section that problem
\eqref{eq:scappmodel} may easily have quite many variables
(columns). One standard approach to solve this type of linear
programming problems is based on iterative generation of only
necessary variables that would lead to an improvement in the objective
function value. This approach is known as \textit{column
  generation}. The main idea is to start with a set of columns
(\textit{column pool}) and form a linear programming problem, called
the \textit{restricted master problem} (RMP). After solving the
restricted master problem, the dual optimal solution is obtained. Then
using this dual solution, a \textit{pricing subproblem} is solved to
find the columns with negative \textit{reduced costs}. These columns
are the only candidates for improving the objective function value
when they are added to the column pool. The next iteration continues
with obtaining the optimal dual solution of the new restricted master
problem with the extended column pool. Again, the pricing subproblem
is solved to find the columns with negative reduced costs. If there is
no such column, then the column generation algorithm terminates.

Suppose that we denote the column pool in the restricted master
problem at iteration $t$ of column generation algorithm by
$\mathcal{J}_t$. Then, we solve the following linear programming
problem:
\begin{equation}
  \label{eq:scappmodelt}
  \begin{array}{lllc}\tag{\texttt{RMP}$(\mathcal{J}_t)$}
    \mbox{minimize\hspace{4mm}} & \sum_{j \in \mathcal{J}_t} (1+w_j) z_j & &\\[2mm]
    \mbox{subject to\hspace{4mm}} & \sum_{j \in \mathcal{J}_t(i)} z_j \geq 1, & i \in \mathcal{I},
                                                              & \hsp (\lambda^t_i)\\[2mm]
                                & z_j \geq 0, & j \in \mathcal{J}_t, &
  \end{array}
\end{equation}
where $\lambda^t_i$ denotes the dual variable corresponding to
constraint $i \in \mathcal{I}$ at iteration $t$, and $\mathcal{J}_t(i)$
stands for the rules in $\mathcal{J}_t$ that cover sample $i$. The
reduced cost of rule $j \in \mathcal{J}_t$ is simply given by
\begin{equation}
  \label{eq:redcost}
  \bar{w}_j = (1+w_j) - \sum_{i \in \mathcal{I}(j)} \lambda^t_i.
\end{equation}
The necessary rules that would improve the objective function value in
the next iteration are the ones with negative reduced costs. In all
this discussion, the key point is, in fact, the pricing subproblem,
which would generate the rules with negative reduced costs. Recall
that the columns in this approach are constructed by the rules in our
setting. Thus, in the remaining part of this section, we shall use the
term ``rule pool'' instead of ``column pool.''

\begin{figure}
  \centering
  \includegraphics[scale = 0.5, trim={2.5cm, 5.5cm, 2.5cm, 5.5cm}]{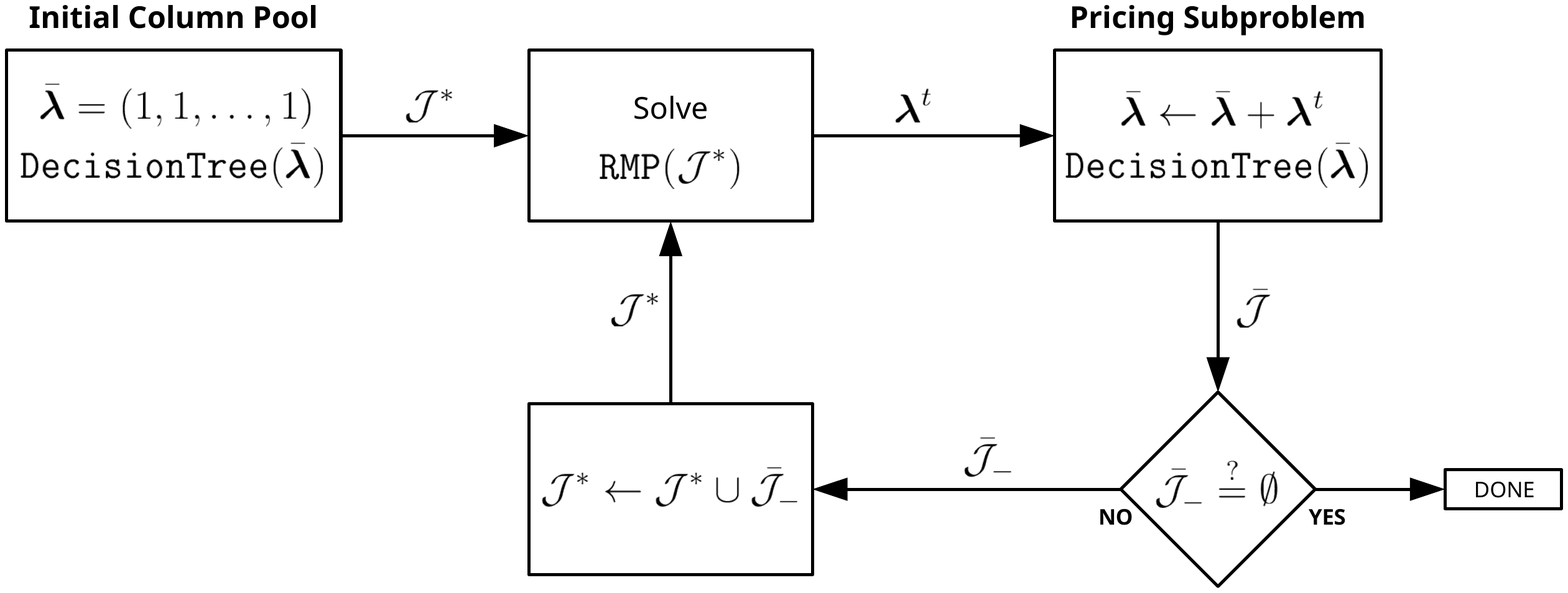}
  \caption{Flowchart of RCBoost}
  \label{fig:flowchart}
\end{figure}

Algorithm \ref{alg:rcboost} shows the training process of the proposed
rule cover boosting (RCBoost) algorithm. The steps of the algorithm is
also given in Figure \ref{fig:flowchart}. Here, \texttt{DecisionTree}
routine plays the role of the pricing subproblem. The parameter of
this routine is a vector, which is used to assign different weights to
the samples. To obtain rules with negative reduced costs, the dual
optimal solution of the restricted master problems is passed as the
sample weight vector (line \ref{ln:psp}). If there are rules with
negative reduced costs (line \ref{ln:nrc}), then these are added to
the rule pool $\mathcal{J}^*$ in line \ref{ln:addJ}. Otherwise, the
algorithm terminates (line \ref{ln:break}). Note that a bound on the
maximum number of RMP calls can also be considered as a
hyperparameter. The resulting set or rules in the final pool
constitute the classifier (line \ref{ln:retJ}). Like in Random Forest
algorithm, a test point is classified using the majority voting system
among all the rules (\texttt{if-then} clauses) that are
satisfied. Note that there is no danger of failing to classify a test
sample like MIRCO, since the initial rule pool is constructed with a
decision tree (line \ref{ln:incp}). The output set of rules
$\mathcal{J}^*$ is then used for classifying out-of-sample point with
majority voting as in \eqref{eq:predict}. That is,
$C(\bm{x}_0, \mathcal{J}^*)$ is the predicted class for $\bm{x}_0$.

\begin{algorithm}
  \SetAlgoLined
  \SetKwInOut{Input}{Input}
  \Input{Training data $(\bm{x}_i, y_i)$, $i \in \mathcal{I}$}
  $\bar{\bm{\lambda}} = (1, 1, \dots, 1)$\;
  $\mathcal{J}^* \leftarrow \text{\texttt{DecisionTree}}(\bar{\bm{\lambda}})$\; \label{ln:incp}
  \For{$t=1, 2, \dots$}{
    $\bm{\lambda}^t \leftarrow \text{\texttt{RMP}}(\mathcal{J}^*)$\;
    $\bar{\bm{\lambda}} \leftarrow \bar{\bm{\lambda}} + \bm{\lambda}^t$\; \label{ln:sumduals}
    $\bar{\mathcal{J}} \leftarrow \text{\texttt{DecisionTree}}(\bar{\bm{\lambda}})$\; \label{ln:psp}
    $\bar{\mathcal{J}}^{-} = \{j \in \bar{\mathcal{J}} ~:~ \bar{w_j} < 0\}$\; \label{ln:nrc}
    \If{$\bar{\mathcal{J}}^{-} = \emptyset$}{
      \textbf{break}\; \label{ln:break}
      }
    $\mathcal{J}^* \leftarrow \mathcal{J}^* \cup \bar{\mathcal{J}}^{-}$\; \label{ln:addJ}
  }
  Return $\mathcal{J}^*$\; \label{ln:retJ}
    \caption{RCBoost}
    \label{alg:rcboost}
\end{algorithm}

As line \ref{ln:sumduals} shows, we accumulate the dual vectors at
every iteration. This is because many of the dual variables quickly
become zero as either the corresponding primal constraints are
inactive or the dual optimal solution becomes degenerate. So, we make
sure with the accumulation that all the sample weights remain
nonzero. We should also note that calling the decision tree algorithm
with (cumulative) sample weights is a proxy to the pricing
subproblem. Because, the actual pricing subproblem is supposed to
search for the rules with negative reduced costs. However, we argue
that using such a proxy is still sensible. We appeal to the theory of
sensitivity analysis in linear programming and the role of sample
weights in boosting algorithms. An optimal dual variable shows the
change in the optimal objective function, when the right-hand side of
this constraint (all ones in our case) is slightly adjusted. Since
each constraint corresponds to the samples in (\ref{eq:scmodel}), the
corresponding optimal dual variables show us the role of these samples
for changing the objective function value. In fact the reduced cost
calculation in \eqref{eq:redcost} also bears the same intuition and
promotes having a rule with low impurity and large dual
variables. Thus, if the samples with large dual values (weights)
belong to the same class, then they are more likely to be classified
within the same leaf (rule) to decrease the impurity. On the other
hand, if these samples belong to different classes, then putting them
into separate leaves also yields a decrease in impurity. This is
precisely what the sample weights try to achieve for boosting methods,
where the weights for the incorrectly classified samples are
increased. For instance, let $w^\lambda_j$ denote the
\textit{weighted} Gini impurity for leaf $j \in \mathcal{J}$ given by
\[
  w^\lambda_j = 1 - \sum_{k \in \mathcal{K}}\left(\frac{\sum_{i \in
        \mathcal{I}(j)} \mathds{1}_k(i)+ \sum_{i \in \mathcal{I}(j)}
      \mathds{1}_k(i)\sum^t_{s=1}\lambda_i^s)}{ |\mathcal{I}(j)| +
      \sum_{i \in \mathcal{I}(j)}\sum^t_{s=1}\lambda_i^s}\right)^2.
\]
where $|\mathcal{I}(j)|$ denotes the cardinality of set
$\mathcal{I}(j)$ and the indicator operator $\mathds{1}_k(i)$ is one
when sample $i$ belongs to class $k$; otherwise, it is zero. This
weighted impurity is minimized whenever the samples with large
accumulated duals from the same class are classified within the same
leaf. This choice is likely to trigger generation of leaves with
negative reduced costs, since the reduced cost value
\eqref{eq:redcost} is also minimized when the same class of samples
(lower Gini impurity) with large duals are covered by rule $j$.

Using sample weights with decision trees also has a clear advantage
from implementation point of view. Almost all existing tools available
for efficiently training decision trees admit weights for the samples
so that boosting algorithms, like the well-known AdaBoost
\cite{freund97}, can make use of this feature. Those
algorithms increase the weights of misclassified samples in an
iterative manner and obtain a sequence of classifiers. Then, these
classifiers are further weighed according to their classification
performances. RCBoost, however, only produces a sequence of rules by
increasing weights and each rule is evaluated by its impurity
contribution to the objective function through its reduced cost. Thus,
the reduced costs play the role of the partial derivatives like in
gradient boosting algorithms. These observations naturally beg for a
discussion about the relationship between our algorithm RCBoost and
the other boosting algorithms. We reserve Section \ref{sec:relation}
to elaborate on this point.

\section{Numerical results}
\label{sec:compstudy}

We next present our computational study with MIRCO and RCBoost
algorithms on 15 datasets with varying characteristics. We have
implemented both algorithms in Python \footnote{(GitHub page) --
  \url{https://github.com/sibirbil/RuleCovering}}. The details of these
datasets are given in Table \ref{tab:datchar}. While most of these
datasets correspond to binary classification problems, there are also
problems with three classes (\texttt{seeds} and \texttt{wine}), six
classes (\texttt{glass}) and eight classes (\texttt{ecoli}). In the
literature, \texttt{mammography}, \texttt{diabetes},
\texttt{oilspill}, \texttt{phoneme}, \texttt{glass} and \texttt{ecoli}
are considered as imbalanced datasets.

MIRCO is compared against Decision Tree (DT) and Random Forest (RF)
algorithms in terms of accuracy and the number of rules
(interpretability). We also compare results of RCBoost (RCB) against
RF, AdaBoost (ADA) and Gradient Boosting (GB). For the latter two
boosting methods, we use decision trees as base estimators. We apply
$10 \times 4$ nested stratified cross validation in all experiments
with the following options for hyperparameter tuning:
$\mathtt{maxdepth} \in \left\{ 5, 10, 20 \right\}$,
$\mathtt{numberoftrees} \in \left\{ 10, 50, 100 \right\}$. We also use
$\mathtt{numberofRMPcalls} \in \left\{ 5, 10, 50, 100, 200 \right\}$
option to set a bound on the number of RMP calls.

\begin{figure}
\centering
\sbox{\measurebox}{%
  \begin{minipage}[b]{.5\textwidth}
  \subfloat
    []
    {\label{fig:accuracy}\includegraphics[scale = 0.5]{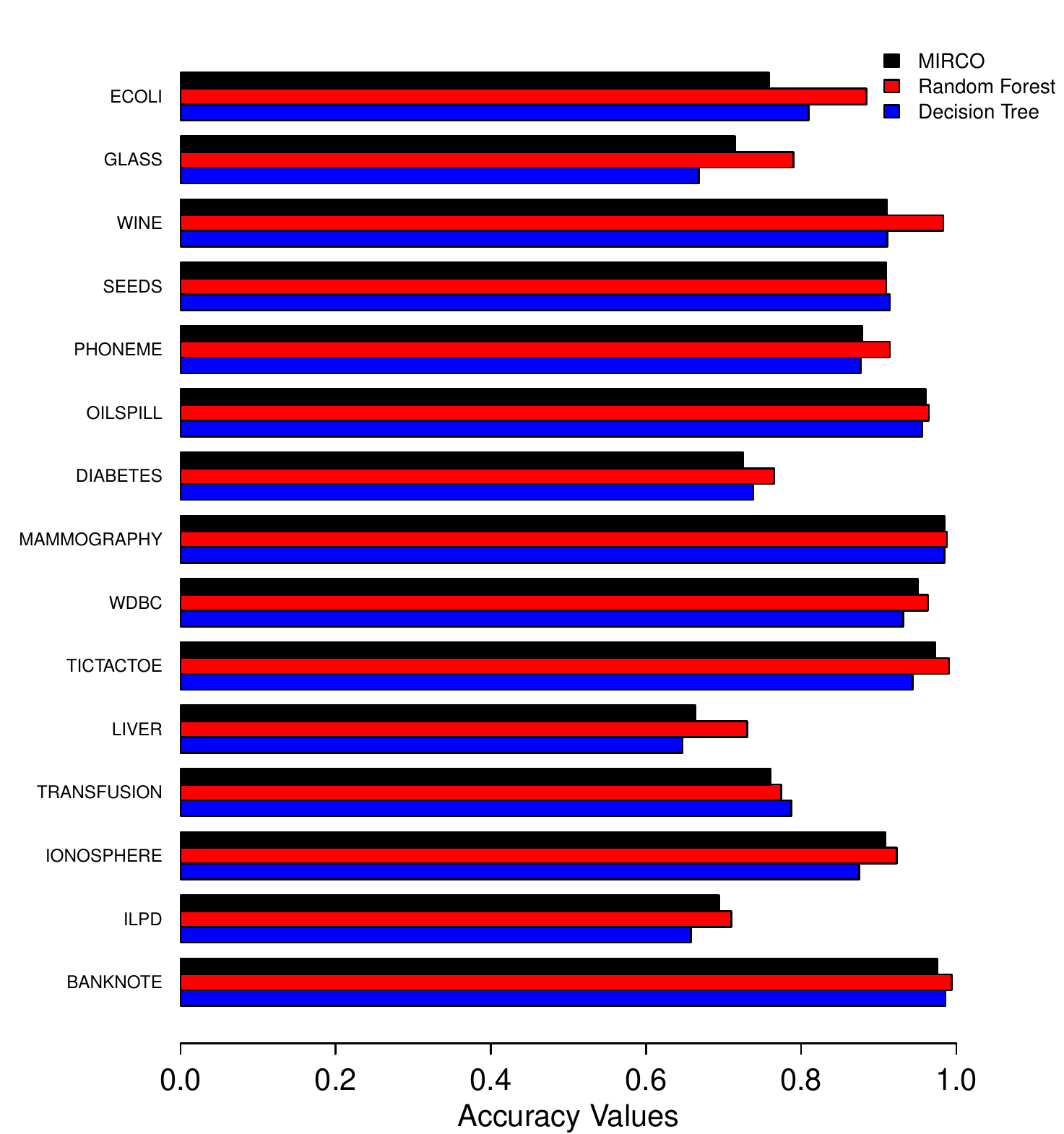}}
  \end{minipage}}
\usebox{\measurebox}\qquad
\begin{minipage}[b][\ht\measurebox][s]{.4\textwidth}
\centering
\subfloat
  []
  {\label{fig:missed}\includegraphics[scale = 0.4]{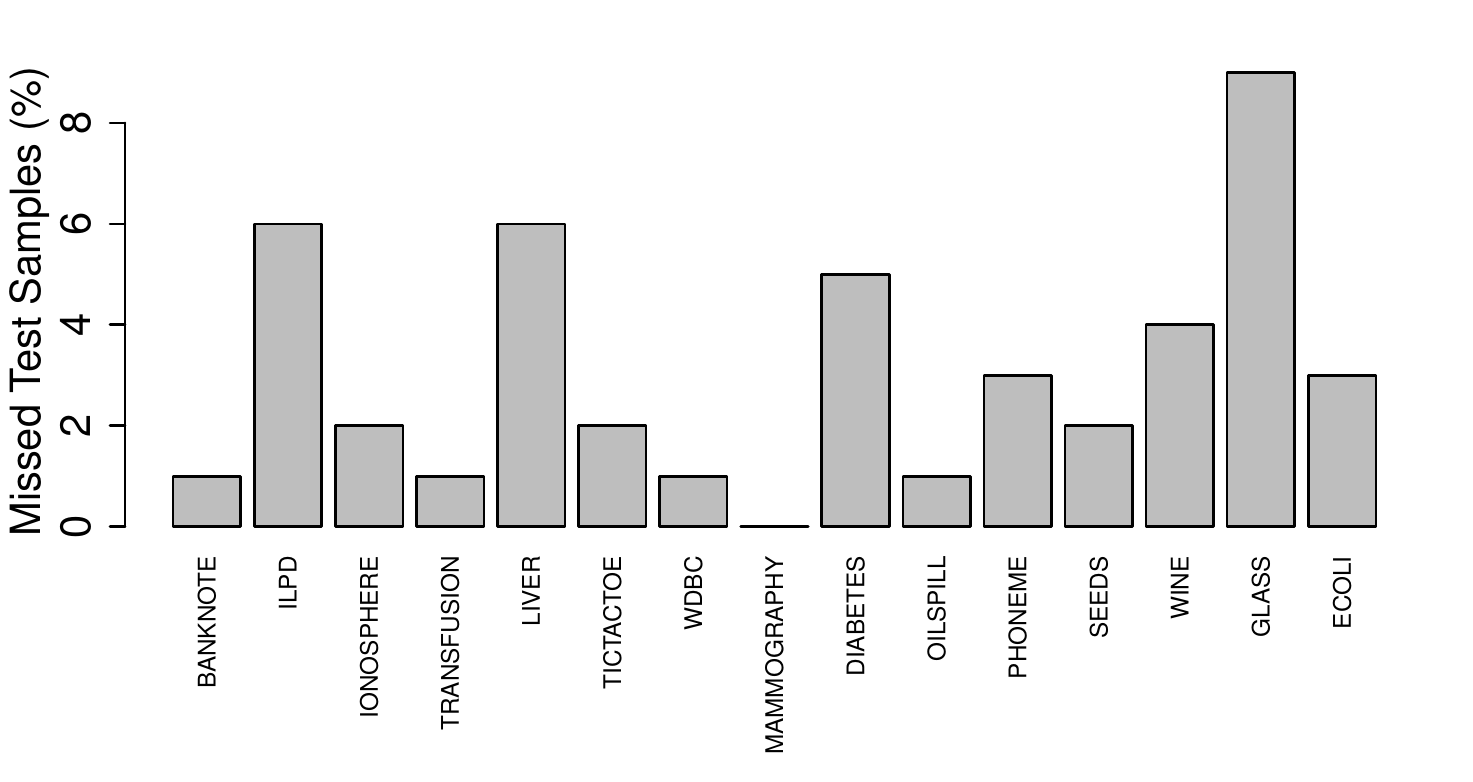}}
\vfill
\subfloat
  []
  {\label{fig:nofrules}\includegraphics[scale = 0.4]{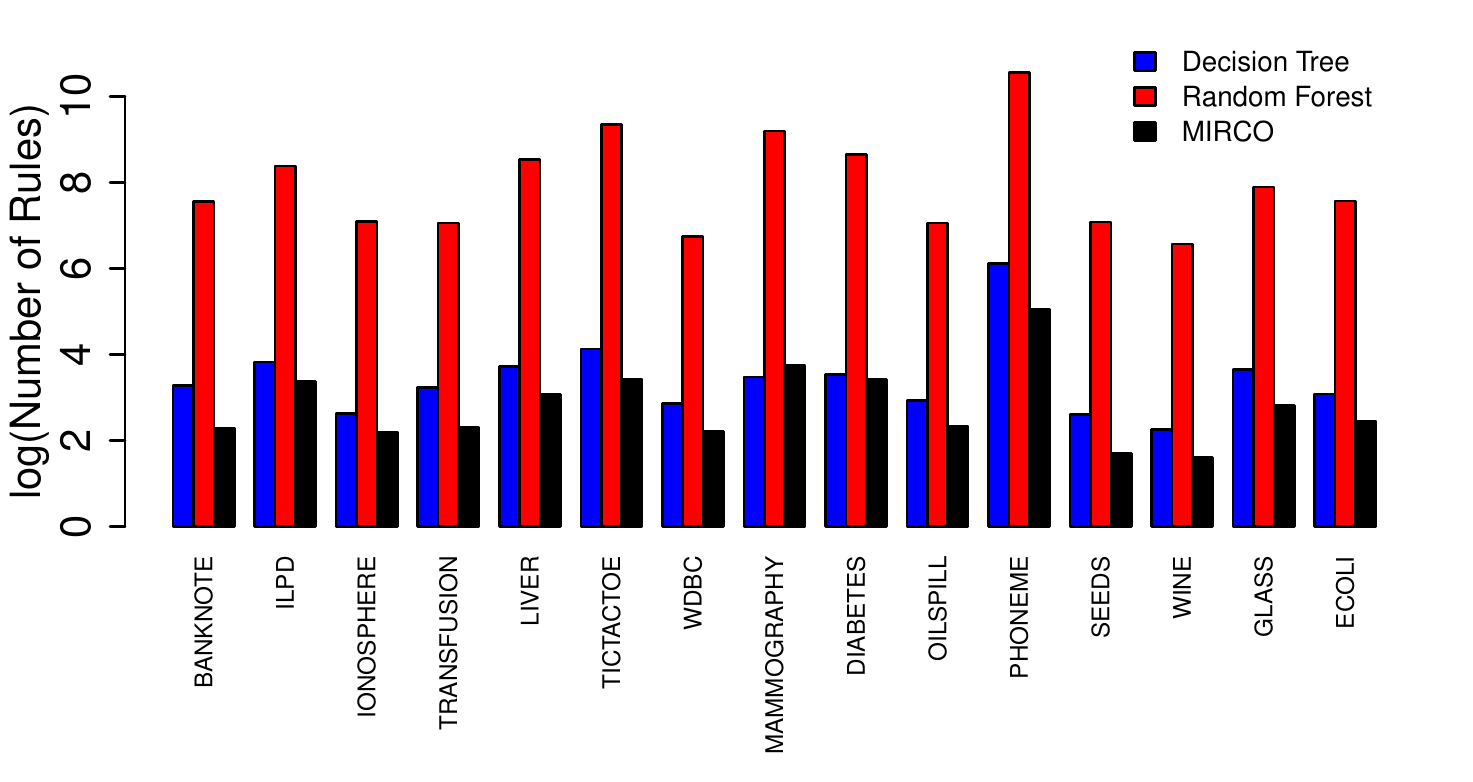}}
\end{minipage}
\caption{(a) Accuracy values for RF, DT, and MIRCO. (b) Average
  percentage of missed samples by MIRCO. (c) Average number of rules
  (in logarithmic scale) generated by RF, DT, and MIRCO.}
\label{fig:mirco}
\end{figure}

We first report our results with DT, RF and MIRCO algorithms in Figure
\ref{fig:mirco}. As the accuracy values (average over 10 replications)
in Figure \ref{fig:accuracy} show, MIRCO performs on par with RF on
most of the datasets (see Table~\ref{tab:mircoresult} for the
details). Recall that MIRCO is not guaranteed to classify all the test
samples. Thus, we also provide the fraction of the test samples missed
by MIRCO (Figure \ref{fig:missed}). However, we still classify such a
sample by applying the rules that have the largest fraction of
accepted clauses. Therefore, the reported comparisons against RF and
DT in Figure \ref{fig:accuracy} are carried on \textit{all}
samples. We observe that the fraction of the missed test samples is
below 6\% for all datasets but one. This shows that MIRCO does not
only approximate the RF in terms of accuracy, but also covers a very
high fraction of the feature space. Figure~\ref{fig:nofrules} depicts
the numbers of rules \textit{in log-scale} to give an indication about
the interpretability. In all but one dataset, MIRCO has generated less
number of rules than DT. Furthermore, the standard deviation figures
in Table~\ref{tab:mircoresult} show that MIRCO shows less variability
than DT in all but two of the datasets.

\begin{figure}
    \centering
    \includegraphics[scale = 0.6]{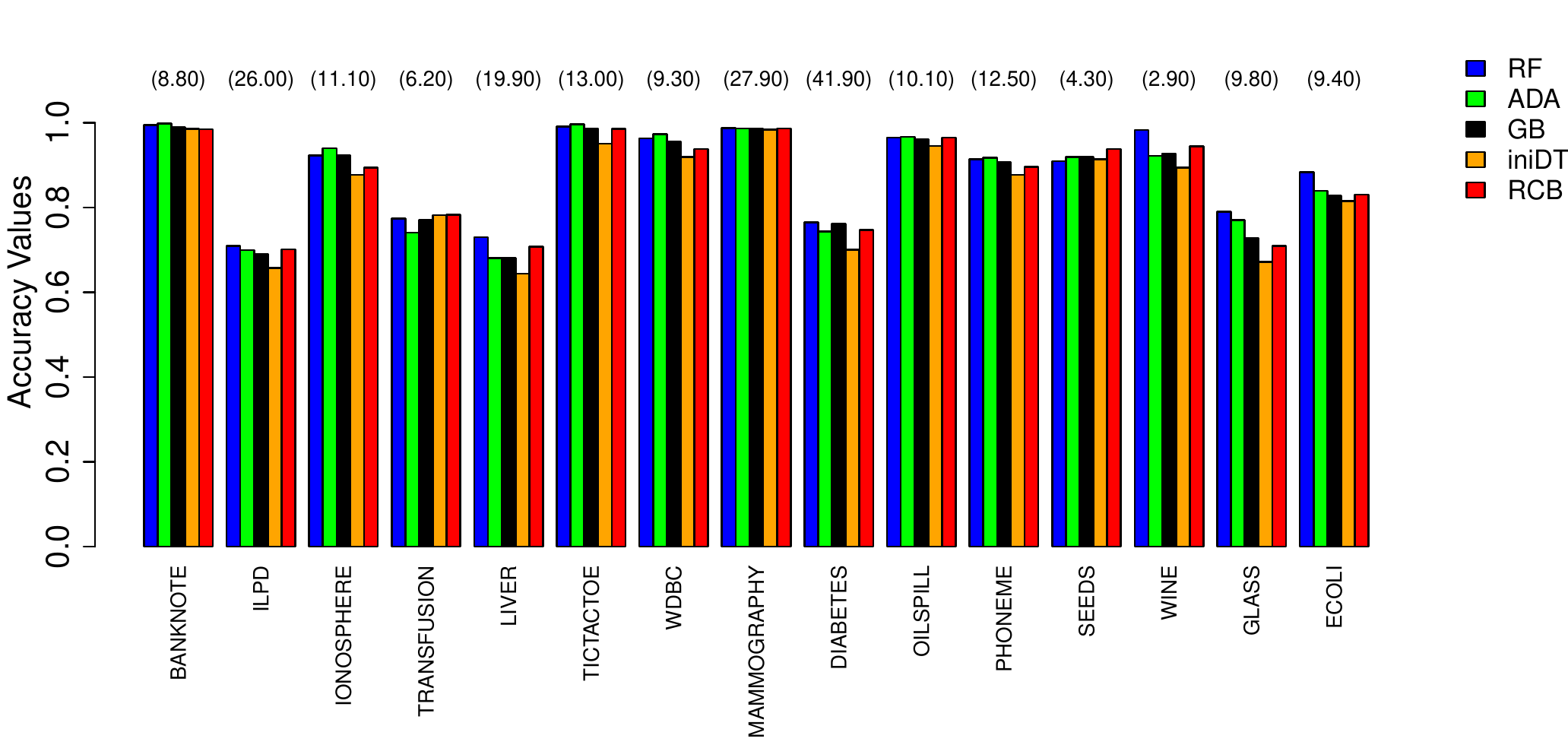}
    \caption{RCBoost results}
    \label{fig:rcboost}
\end{figure}

To compare RCB, we present the average accuracy values of RF, ADA, GB
in Figure~\ref{fig:rcboost} (see Table~\ref{tab:rcboostresult} for the
details). We also report the results with the initial Decision Tree
(iniDT) used in RCB algorithm (see line~\ref{ln:incp} in
Algorithm~\ref{alg:rcboost}) so that we can observe how rule
generation improves performance. The number in parentheses above each
dataset shows the average number of RMP calls by RCB. These results
demosntrate that RCB exhibits a quite competitive performance. In
particular, RCB outperforms both boosting algorithms, ADA and GB in
four problems. One critical point here is that RCB makes only a small
number of RMP calls. In all but one datasets, RCB improves the
performance of the base decision tree, iniDT by using the proposed
rule generation.

\clearpage

\section{Conclusion}
\label{sec:conclusion}

We have proposed two algorithms for interpretation and boosting. Both
algorithms are based on mathematical programming approaches for
covering the rules in decision trees. The first algorithm is used for
interpreting a model trained with the Random Forest algorithm. The
second algorithm focuses on the idea of generating necessary rules to
boost an initial model trained with the Decision Tree algorithm. Both
algorithms are quite straightforward to implement within the existing
packages. Even with our first implementation attempt, we have obtained
promising results. The first algorithm has successfully extracted much
less number of rules than the random forest model leading to better
interpretability. Moreover, we have observed that the accuracy levels
obtained with the extracted rules of our first algorithm closely follow
the accuracy levels of the trained random forest model. When compared
against other boosting algorithms, our second algorithm has shown a
competitive performance. As a new boosting algorithm, we have also
provided a discussion about the relationship between our approach and
other boosting algorithms. The idea of selecting rules for
interpretation or boosting leads to a plethora of extensions that
could be further studied. Therefore, we have prepared a separate
supplementary discussion on extensions in Section
\ref{sec:extensions}. This section complements our discussion as well
as lists several future research paths.

\small

\clearpage

\bibliography{ref}

\clearpage

\renewcommand\thesection{S.\arabic{section}}
\setcounter{section}{0}
\renewcommand\thetable{S.\arabic{table}}
\setcounter{table}{0}
\renewcommand{\thealgocf}{S.\arabic{algocf}}
\setcounter{algocf}{0}

\begin{center}
  {\Large Supplementary Material for \\[2mm]
``Rule Covering for Interpretation and Boosting''}
\end{center}

\bigskip

Note that most of the cross references in this supplementary file
refer to the original manuscript. Therefore, clicking on those
references may not take you to the desired location.

\section{Chvatál's greedy algorithm}
\label{sec:greedy}

We restate the well-known heuristic of Chvatál in our own
notation. Recall that the sets $\mathcal{I}$ and $\mathcal{J}$ stand
for the set of samples and the set of rules, respectively. Algorithm
\ref{alg:greedy} gives the steps of the greedy heuristic, where
$\mathcal{I}_{\mathcal{R}}(j)$ represents those samples that are in
$\mathcal{R}$ covered by rule $j \in \mathcal{J}$.  In line
\ref{ln:ratio}, the cost of selecting rule $j$ is $1 + w_j$, and the
operation $|\mathcal{I}_{\mathcal{R}}(j)|$ denotes the cardinality of
the set $\mathcal{I}_{\mathcal{R}}(j)$. Removal of the redundant rules
in line \ref{ln:redundant} is done in two steps: First, the rules in
$\bar{\mathcal{J}}$ are sorted in ascending order of their costs in a
list. Then, the rules at the end of this list are removed as long as
the remaining rules cover all the samples.

\begin{algorithm}
  \SetAlgoLined
  \KwResult{Set of rules $\hat{\mathcal{J}}$}

  $\mathcal{R} = \mathcal{I}$, $\hat{\mathcal{J}} = \emptyset$\;

  \While{$\mathcal{R} \neq \emptyset$}{

    $j^* = \arg\underset{j \in \mathcal{J} \backslash \hat{\mathcal{J}}}{\min}\left\{
      \frac{1 + w_j}{|\mathcal{I}_{\mathcal{R}}(j)|} :
      \mathcal{I}_{\mathcal{R}}(j)\neq\emptyset\right\}$\; \label{ln:ratio}

    $\mathcal{R} \leftarrow \mathcal{R} \backslash \mathcal{I}_{\mathcal{R}}(j^*)$\;

    $\hat{\mathcal{J}} \leftarrow \hat{\mathcal{J}} \cup j^*$\;
  }

  Remove redundant columns from $\hat{\mathcal{J}}$\; \label{ln:redundant}

  Return $\hat{\mathcal{J}}$\;
    \caption{Greedy}
    \label{alg:greedy}
\end{algorithm}

\bigskip

\section{Relation to other boosting algorithms}
\label{sec:relation}

The use of sample weights and reduced costs shows a similar trait
among RCBoost and other boosting algorithms. The reduced cost of rule
$j$ can be considered as the partial derivative of the objective
function with respect to the corresponding variable $z_j$. Thus, rules
that have negative reduced costs are added to the rule pool. This
resembles a descent iteration.  To establish the relationship between
RCBoost and other boosting methods from the gradient descent point of
view \citep{mason99}, we first define the Lagrangean function
\[
  \mathcal{L}(\mathbf{z}, \bm{\lambda}) = \sum_{j \in
    \mathcal{J}}\big(1 + w_j\big)z_j + \sum_{i \in
    \mathcal{I}}\lambda_i \big(1-\sum_{j \in \mathcal{J}(i)}z_j\big),
\]
where $\mathbf{z}$ and $\bm{\lambda}$ denote the primal and dual
vectors. Suppose that the optimal solution pair for
\eqref{eq:scappmodelt} is given by $\mathbf{z}^t$ and
$\bm{\lambda}^t$. Then, we have
\[
  \begin{array}{rl}
  \mathcal{L}((\mathbf{z}^t, \mathbf{z}^c), \bm{\lambda}^t) & = \sum_{i
    \in \mathcal{I}}\lambda^t_i + \sum_{j \in
    \mathcal{J}_t}\bar{w}_jz^t_j + \sum_{j \in
    \mathcal{J}/\mathcal{J}_t}\bar{w}_jz^t_j \\[2mm]
   &  = \sum_{i \in
    \mathcal{I}}\lambda^t_i + \sum_{j \in
    \mathcal{J}/\mathcal{J}_t}\bar{w}_jz^c_j,
  \end{array}
\]
where $\bar{w}_j$ is the reduced cost of rule $j$ as in
\eqref{eq:redcost} and the tuple $(\mathbf{z}^t, \mathbf{z}^c)$ shows
the variables in $\mathcal{J}_t$ and $\mathcal{J}/\mathcal{J}_t$,
respectively. The last equality is a direct consequence of optimality
conditions, since $\mathbf{z}^t$ and $\bm{\lambda}^t$ are
complementary solution vectors. Thus, taking the steepest descent step
for $j \in \mathcal{J}/\mathcal{J}_t$ simply implies involving those
rules with negative reduced costs in the next iteration. From this
perspective, RCBoost is similar to gradient boosting
algorithms. Nonetheless, unlike those algorithms RCBoost does not fit
a weak-classifier but instead obtains a larger set $\mathcal{J}_{t+1}$
by adding a subset of rules with negative reduced costs (line
\ref{ln:addJ} of Algorithm \ref{alg:rcboost}). Then, solving the
linear programming problem over $\mathcal{J}_{t+1}$ yields the
parameters $\mathbf{z}^{t+1}$ and $\bm{\lambda}^{t+1}$. Suppose we
denote the optimal objective function of \eqref{eq:scappmodelt} by
$\Phi(\mathcal{J}_t)$, which is the total impurity that we use to
evaluate the training loss of our classifier. With each iteration we
improve this loss, since
$\Phi(\mathcal{J}_{t}) \geq \Phi(\mathcal{J}_{t+1}) $ due to our
linear programming formulation. If we further define the subset

\[
  \partial \mathcal{J}_t \subseteq \{j \in \mathcal{J}: \bar{w}_j < 0\},
\]

then $\Phi(\mathcal{J}_{t+1}) = \Phi(\mathcal{J}_t \cup \partial \mathcal{J}_t)$.

There has also been a focus on the margin maximization idea to discuss
the generalization performance of boosting methods \citep{mason00,
  schapire98}. Here, one of the fundamental points is to observe that
a sample with a large margin is likely to be classified
correctly. Thus, margin maximization is about assigning larger weights
to those samples with small margins. RCBoost also implicitly aims at
obtaining large margins through selection of rules with less
impurities (classification errors decrease). In other words,
minimizing the objective function of the restricted master problem
promotes large margins. Formally, we can rewrite the optimal objective
function value of \eqref{eq:scappmodelt} as
\[
  \sum_{j \in \mathcal{J}_t} (1+w_j) z^t_j = \sum_{i\in\mathcal{I}}\sum_{j \in \mathcal{J}_t(i)}
  \mu_j z^t_j = \sum_{i\in\mathcal{I}}\lambda_i^t,
\]
where $\mu_j = \frac{1+w_j}{|\mathcal{I}(j)|}$, $j \in
\mathcal{J}$. Using complementary slackness condition, we know that
$\lambda_i^t > 0$ only if $\sum_{j \in \mathcal{J}_t(i)} z_j^t =
1$. Thus, $\sum_{j \in \mathcal{J}_t(i)} \mu_j z^t_j$ is the convex
combination of margins obtained with different rules for sample
$i$. The larger the obtained impurity, the smaller the corresponding
margin. Like other boosting methods, RCBoost also assigns larger
weights to those samples with smaller margins.

\section{Some extensions}
\label{sec:extensions}

We have used classification for our presentation. Nonetheless, both
proposed algorithms can also be adjusted to solve regression
problems. To interpret random forest regressors with MIRCO, the first
step is to use a proper criterion, like mean squared error, for
evaluating $w_j$, $j \in \mathcal{J}$ values. Then, MIRCO can be used
as it is given in Algorithm \ref{alg:mirco}. The same criteria can
also be used in RCBoost to construct the restricted master
problem. When a regression problem is solved, one needs to pay
attention to two potential caveats: (i) The two components of the
objective function coefficients in optimization model
\eqref{eq:scmodel} or \eqref{eq:scappmodel} may not be balanced well,
since the criteria values $w_j$, $j \in \mathcal{J}$ are not bounded
above by one. This could be overcome by scaling the criteria
values. If scaling is done with a multiplier, then this multiplier
becomes a new hyperparameter which may require tuning. (ii) Regression
trees tend to have too many leaves. Even after applying MIRCO, the
cardinality of the resulting set of rules $\hat{\mathcal{J}} $ may be
quite large, and hence, interpreting the random forest regressor may
be difficult. Similarly, RCBoost may make too many calls to the
restricted master problem and end up with a large number of rules in
the final rule pool $\mathcal{J}^*$.

Many boosting algorithms use decision trees as their weak
estimators. Thus, it seems that MIRCO can also be used with those
algorithms to interpret their results. However, one needs to pay
attention to the fact that boosting algorithms assign different
weights to their weak estimators and MIRCO does not take these weights
into account. Therefore, if rule covering is used for interpretation
of boosting algorithms, Algorithm \ref{alg:mirco} should be adjusted
to incorporate the estimator weights.

Recall that we use a decision tree with sample weights in RCBoost as a
proxy to an actual pricing subproblem (line \ref{ln:psp}, Algorithm
\ref{alg:rcboost}). An alternative could be using the impurity and the
negative reduced cost together while constructing the decision
tree. In that case, splitting can be done from non-dominated
bi-criteria values; \textit{i.e.,} impurity and negative reduced cost
values.

Clearly, we can also use a random forest or any ensemble of trees with
sample weights as a proxy to the pricing subproblem. Even further, one
may even model a separate combinatorial problem to generate many
different rules. At this point, it is important to keep in mind that
considering \textit{all} possible rules with negative reduced costs
could prove to be a daunting job.

In its current stage, RCBoost adds all the rules with negative reduced
costs to the rule pool. An alternative approach could be to include
only the rule with the most negative reduced cost. Along the same
vein, we may also remove those rules that have high (positive) reduced
costs from the column pool. Actually, these are typical approaches in
column generation to keep the size of the column pool manageable. As
the size of the rule pool becomes large, RCBoost also becomes less
interpretable. Fortunately, one can easily apply MIRCO or other rule
extraction approaches from the literature such as the RuleFit
algorithm \cite{friedman2008predictive} to a trained RCBoost. In that
case, Algorithm \ref{alg:mirco} can be applied starting from line
\ref{ln:ei} with $\mathcal{J} \leftarrow \mathcal{J}^*$. As a final
remark, we note that RCBoost treats the rules equally when a test
point is classified with the satisfied rules in $\mathcal{J}^*$. An
alternative could be giving different weights to different
rules. These rule weights could be evaluated by making use of the
optimal values of $z_j$, $j \in \mathcal{J}^*$ obtained from the final
restricted master problem. However, all these changes should be
carefully contemplated, since the resulting rule pool after shrinking
may not necessarily cover the feature space as we have discussed in
Section \ref{sec:mirco} (see also Figure \ref{fig:misses}).

\clearpage

\section{Details of numerical results}
\label{sec:datasets}

\begin{table}[h]
  \caption{Characteristics of the datasets}
  \label{tab:datchar}
  \centering
  \begin{tabular}{lccc}
    \toprule
     & \multicolumn{1}{c}{Number of} & \multicolumn{1}{c}{Number of} & \multicolumn{1}{c}{Number of} \\
    Dataset & Features & Classes & Samples \\
    \midrule
    banknote & 4 & 2 & 1372 \\
    ILDP & 10 & 2 & 583 \\
    ionosphere & 34 & 2 & 351 \\
    transfusion & 4 & 2 & 748 \\
    liver & 6 & 2 & 345  \\
    tic-tac-toe & 9 & 2 & 958 \\
    WDBC & 30 & 2 & 569 \\
    mammography & 6 & 2 & 11183 \\
    diabetes & 8 & 2 & 768 \\
    oilspill & 48 & 2 & 937 \\
    phoneme & 5 & 2 & 5427 \\
    seeds & 7 & 3 & 210 \\
    wine & 13 & 3 & 178 \\
    glass & 10 & 6 & 214 \\
    ecoli & 7 & 8 & 336 \\
    \bottomrule
  \end{tabular}
\end{table}

\begin{table}[h]
  \caption{MIRCO results}
  \label{tab:mircoresult}
  \centering
  \begin{tabular}{lcccc}
    \toprule
     & \multicolumn{1}{c}{DT} & \multicolumn{1}{c}{RF} & \multicolumn{2}{c}{MIRCO} \\
    \cmidrule(r){4-5}
    Dataset & Accuracy & Accuracy & Accuracy & \% of Missed Pts\\
    \midrule
banknote	&	0.99 (0.02)	&	0.99 (0.01)	&	0.98 (0.01)	&	0.01 (0.01)	\\
ILDP	&	0.66 (0.07)	&	0.71 (0.06)	&	0.69 (0.03)	&	0.06 (0.04)	\\
ionosphere	&	0.87 (0.07)	&	0.92 (0.03)	&	0.91 (0.04)	&	0.02 (0.03)	\\
transfusion	&	0.79 (0.04)	&	0.77 (0.04)	&	0.76 (0.01)	&	0.01 (0.01)	\\
liver	&	0.65 (0.09)	&	0.73 (0.06)	&	0.66 (0.06)	&	0.06 (0.05)	\\
tic-tac-toe	&	0.94 (0.02)	&	0.99 (0.01)	&	0.97 (0.02)	&	0.02 (0.02)	\\
WDBC	&	0.93 (0.03)	&	0.96 (0.02)	&	0.95 (0.03)	&	0.01 (0.02)	\\
mammography	&	0.98 (0.00)	&	0.99 (0.00)	&	0.98 (0.00)	&	0.00 (0.00)	\\
diabetes	&	0.74 (0.04)	&	0.77 (0.04)	&	0.73 (0.04)	&	0.05 (0.04)	\\
oilspill	&	0.96 (0.02)	&	0.96 (0.01)	&	0.96 (0.01)	&	0.01 (0.01)	\\
phoneme	&	0.88 (0.01)	&	0.91 (0.01)	&	0.88 (0.02)	&	0.03 (0.01)	\\
seeds	&	0.91 (0.06)	&	0.91 (0.09)	&	0.91 (0.05)	&	0.02 (0.03)	\\
wine	&	0.91 (0.07)	&	0.98 (0.03)	&	0.91 (0.07)	&	0.04 (0.04)	\\
glass	&	0.67 (0.05)	&	0.79 (0.07)	&	0.71 (0.07)	&	0.09 (0.06)	\\
ecoli	&	0.81 (0.09)	&	0.88 (0.06)	&	0.76 (0.08)	&	0.03 (0.03)	\\

    \bottomrule
    \multicolumn{4}{l}{$^*$ Numbers in parantheses show the standard deviations}
  \end{tabular}
\end{table}

\begin{table}[h]
  \caption{RCBoost results}
  \label{tab:rcboostresult}
  \centering
  \begin{tabular}{lccccc}
    \toprule
     & \multicolumn{1}{c}{RF} & \multicolumn{1}{c}{ADA} & \multicolumn{1}{c}{GB} & \multicolumn{1}{c}{iniDT} & \multicolumn{1}{c}{RCB} \\
    Dataset & Accuracy & Accuracy & Accuracy & Accuracy & Accuracy \\
    \midrule
banknote	&	0.99 (0.01)	&	1.00 (0.00)	&	0.99 (0.01)	&	0.99 (0.01)	&	0.98 (0.01)	\\
ILDP	&	0.71 (0.06)	&	0.70 (0.06)	&	0.69 (0.03)	&	0.66 (0.07)	&	0.70 (0.04)	\\
ionosphere	&	0.92 (0.03)	&	0.94 (0.05)	&	0.92 (0.04)	&	0.88 (0.03)	&	0.89 (0.05)	\\
transfusion	&	0.77 (0.04)	&	0.74 (0.04)	&	0.77 (0.03)	&	0.78 (0.03)	&	0.78 (0.04)	\\
liver	&	0.73 (0.06)	&	0.68 (0.07)	&	0.68 (0.06)	&	0.64 (0.10)	&	0.71 (0.05)	\\
tic-tac-toe	&	0.99 (0.01)	&	1.00 (0.01)	&	0.99 (0.01)	&	0.95 (0.02)	&	0.99 (0.01)	\\
WDBC	&	0.96 (0.02)	&	0.97 (0.03)	&	0.96 (0.03)	&	0.92 (0.03)	&	0.94 (0.02)	\\
mammography	&	0.99 (0.00)	&	0.99 (0.00)	&	0.99 (0.00)	&	0.98 (0.00)	&	0.99 (0.00)	\\
diabetes	&	0.77 (0.04)	&	0.74 (0.04)	&	0.76 (0.04)	&	0.70 (0.06)	&	0.75 (0.03)	\\
oilspill	&	0.96 (0.01)	&	0.97 (0.01)	&	0.96 (0.02)	&	0.95 (0.02)	&	0.96 (0.02)	\\
phoneme	&	0.91 (0.01)	&	0.92 (0.01)	&	0.91 (0.01)	&	0.88 (0.01)	&	0.90 (0.01)	\\
seeds	&	0.91 (0.09)	&	0.92 (0.07)	&	0.92 (0.07)	&	0.91 (0.09)	&	0.94 (0.06)	\\
wine	&	0.98 (0.03)	&	0.92 (0.08)	&	0.93 (0.05)	&	0.89 (0.07)	&	0.94 (0.04)	\\
glass	&	0.79 (0.07)	&	0.77 (0.08)	&	0.73 (0.10)	&	0.67 (0.11)	&	0.71 (0.09)	\\
ecoli	&	0.88 (0.06)	&	0.84 (0.07)	&	0.83 (0.06)	&	0.82 (0.09)	&	0.83 (0.05)	\\
    \bottomrule
    \multicolumn{4}{l}{$^*$ Numbers in parantheses show the standard deviations}
  \end{tabular}
\end{table}

\end{document}